%% file: main.tex
\documentclass[runningheads]{llncs}
\usepackage{float}
\usepackage{comment}
\usepackage{graphicx}
\usepackage{amsmath, bm}
\usepackage{amssymb}
\usepackage{booktabs}
\usepackage{threeparttable,booktabs}
\usepackage{longtable}
\usepackage[hang,flushmargin]{footmisc}
\usepackage{multirow}
\usepackage[pagebackref,breaklinks,colorlinks]{hyperref}
\usepackage{utfsym}
\usepackage{subfigure}
\usepackage{colortbl}
\usepackage{xcolor}
\definecolor{lightergray}{gray}{0.9}

\newcommand{\model}[0]{{\textcolor{black}{PMC-CLIP}}}
\newcommand{\dataset}[0]{{\textcolor{black}{PMC-OA}}}
\newcommand{\datasetbeta}[0]{{\textcolor{black}{PMC-OA Beta}}} 

\begin{document}

\title{PMC-CLIP: Contrastive Language-Image Pre-training using Biomedical Documents}
%
%
\author{Weixiong Lin\inst{1, *} \and
Ziheng Zhao\inst{1, *} \and
Xiaoman Zhang \inst{1,2} \and
Chaoyi Wu \inst{1,2} \and
Ya Zhang \inst{1,2} \and
Yanfeng Wang \inst{1,2} \and
Weidi Xie \inst{1,2,\dag}
}
%
%
%
%
\institute{Cooperative Medianet Innovation Center, Shanghai Jiao Tong University, Shanghai, China \\
\and Shanghai AI Laboratory, Shanghai, China
\email{\{wx\_lin,Zhao\_Ziheng,xm99sjtu,wtzxxxwcy02,ya\_zhang,wangyanfeng,weidi\}@sjtu.edu.cn}
}
\maketitle              
\begin{abstract}
Foundation models trained on large-scale dataset gain a recent surge in CV and NLP.
In contrast, development in biomedical domain lags far behind due to data scarcity.
To address this issue, we build and release \dataset{}, a biomedical dataset with 1.6M image-caption pairs collected from PubMedCentral's OpenAccess subset, 
which is 8 times larger than before.
\dataset{} covers diverse modalities or diseases, with majority of the image-caption samples aligned at finer-grained level, {\em i.e.}, subfigure and subcaption.
While pretraining a CLIP-style model on \dataset{}, our model named \model{} achieves state-of-the-art results on various downstream tasks, 
including image-text retrieval on ROCO, MedMNIST image classification, Medical VQA, {\em i.e.} +8.1\% R@10 on image-text retrieval, +3.9\% accuracy on image classification.

\keywords{Foundation Model  \and Multimodal Dataset \and Vision-Language Pretraining.}
\end{abstract}

\renewcommand{\thefootnote}{}
\footnotetext{\dag: Corresponding author.~~ *: These authors contribute equally to this work.}

\input{content/01-Introduction.tex}
\input{content/02-DatasetCollection.tex}

\input{content/03-Method.tex}

\input{content/04-Experiment.tex}

\input{content/05-Result.tex}

\input{content/06-Conclusion.tex}

\newpage
\bibliographystyle{ieee_fullname}
\bibliography{egbib}

\end{document}

%% file: content/01-Introduction.tex
\section{Introduction}


In the recent literature, 
development of foundational models has been the main driving force in artificial intelligence, for example, large language models~\cite{ramesh2022hierarchical,ding2022cogview2,brown2020language,ouyang2022training} trained with either autoregressive prediction or masked token inpainting,
and computer vision models~\cite{li2021align,radford2021learning,wang2022image} 
trained by contrasting visual-language features.
In contrast, development in the biomedical domain lags far behind due to limitations of data availability from two aspects, 
(i) the expertise required for annotation, (ii) privacy concerns. 
This paper presents our preliminary study for constructing a {\bf large-scale, 
high-quality, image-text} biomedical dataset using publicly available scientific papers, with {\bf minimal manual efforts} involved. 

\begin{figure}[t]
\centering
\includegraphics[width=0.99\textwidth]{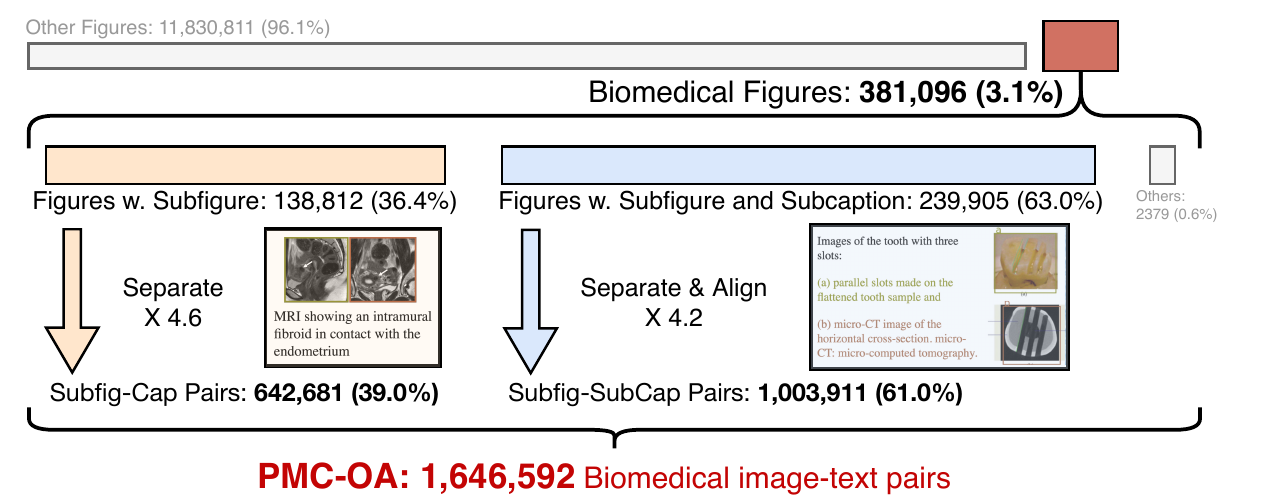}
\vspace{-0.3cm}
\caption{Statistics over the pipeline and the collected \dataset.}
\vspace{-0.6cm}
\label{fig:Statis_Visualization}
\vspace{-3pt}
\end{figure}

In particular, we crawl figures and corresponding captions from scientific documents on PubMed Central, which is a free full-text archive of biomedical and life sciences journal literature at the U.S. National Institutes of Health's National Library of Medicine (NIH/NLM)~\cite{roberts2001pubmed}. This brings two benefits: (i) the contents in publications are generally well-annotated and examined by experts, (ii) the figures have been well-anonymized and de-identified. In the literature, we are clearly not the first to construct biomedical datasets in such a manner, 
however, existing datasets suffer from certain limitations from today's standard.
For example, as a pioneering work,
ROCO~\cite{pelka2018roco} was constructed long time ago with only 81k radiology images.
MedICAT~\cite{subramanian2020medicat} contains 217k images, 
but are mostly consisted of compound figures. 
In this work, we tackle the above-mentioned limitations by introducing an automatic pipeline to generate dataset with subfigure-subcaption correspondence from scientific documents, consisting of three major stages:
medical figure collection, subfigure separation, subcaption separation $\&$ alignment.
The final dataset, \dataset, consisting of 1.65M image-text pairs, 
Fig.~\ref{fig:Statis_Visualization} and Fig.~\ref{fig:radar statistics}.


Along with the constructed dataset, 
we train a CLIP-style vision-language model for biomedical domain, termed as \model. The model is trained on \dataset~with standard image-text contrastive (ITC) loss, 
and to encourage the joint interaction of image and text, 
masked language modeling (MLM) is also applied. 
We evaluate the pre-trained model on several downstream tasks, 
including medical image-text retrieval, medical image classification,
and medical visual question answering~(VQA). \model~achieves state-of-the-art performance on various downstream tasks, surpassing previous methods significantly.

Overall, in this paper, we make the following contributions:
\textbf{First}, we propose an automatic pipeline to construct high-quality image-text biomedical datasets from scientific papers,
and construct an image-caption dataset via the proposed pipeline, named \dataset{}, which is 8$\times$ larger than before. 
With the proposed pipeline, the dataset can be continuously updated.
\textbf{Second}, we pre-train a vision-language model, termed as \model{}, on the constructed image-caption dataset, to serve as a foundation model for the biomedical domain.
\textbf{Third}, we conduct thorough experiments on various tasks~(retrieval, classification, and VQA),
and obtain SOTA performance on most downstream datasets,
demonstrating the superiority of \dataset{} and the potential of the foundation model~\model.
The dataset and pre-trained model will be made available to the community.

%% file: content/02-DatasetCollection.tex
\section{The \dataset~Dataset}

In this section, we start by describing the dataset collection procedure in Sec.~\ref{sec:dataset_collection},
followed by a brief overview of \dataset~in Sec.~\ref{sec:dataset_overview}.

\begin{figure}[t] 
\centering 
\includegraphics[width=\textwidth]{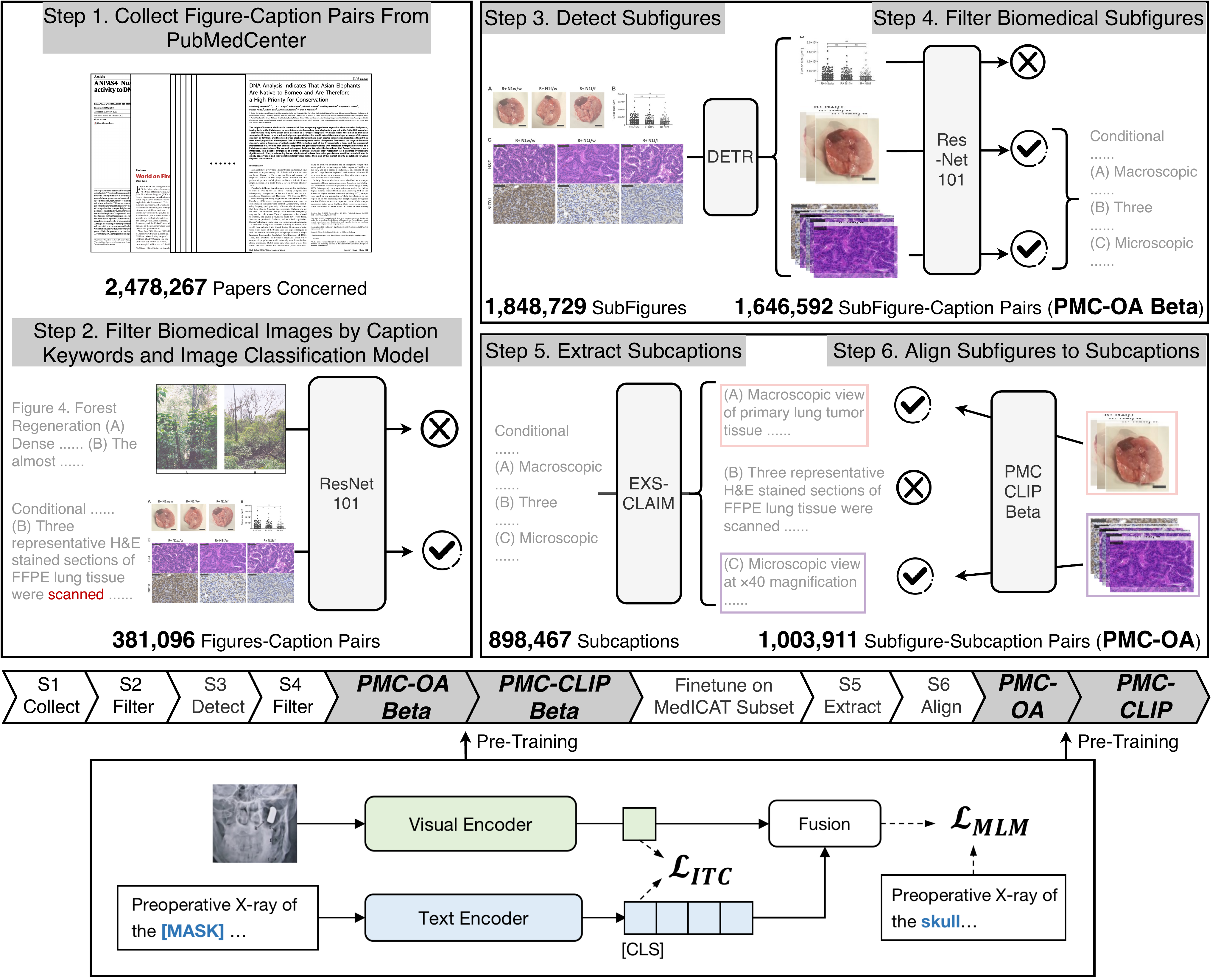}
\caption{The proposed pipeline to collect PMC-OA~(upper) and the architecture of PMC-CLIP~(bottom).}
\label{fig:workflow}
\end{figure}

\subsection{Dataset Collection}
\label{sec:dataset_collection}
In this section, we detail the propose pipeline for creating \dataset, 
a large-scale dataset that contains 1.65M image-text pairs.
The whole procedure consists of three major stages: 
(i) medical figure collection, 
(ii) subfigure separation, 
(iii) subcaption separation~$\&$~alignment, 
as summarised in Fig.~\ref{fig:workflow}.

\vspace{3pt}
\noindent \textbf{Medical Figure Collection~(Step 1\&2 in Fig.~\ref{fig:workflow}).} We first extract figures and captions from PubMed Central, 
by the time of 2022-09-16, 2,478,267 available papers are covered and 12,211,907 figure-caption pairs are extracted. To derive medical figures, we follow the same procedure as in~\cite{subramanian2020medicat}: 
{\em first}, a set of medical keywords\footnote{Follow Class T060 ``\textit{Diagnostic Procedure}'' defined in UMLS~\cite{bodenreider2004unified}} are pre-defined to filter the captions, deleting the figure-caption pair with no keywords appearing in the caption; {\em second}, a ResNet-101~\cite{he2016deep} trained on DocFigure~\cite{jobin2019docfigure} for scientific figure classification is applied to classify the remaining figures into 28 categories. We keep the figures with ``Medical'' class prediction scores in Top 4 of 28 categories, ending up with 381,096 medical figures.

\vspace{3pt}
\noindent \textbf{Subfigure Separation~(Step 3\&4 in Fig.~\ref{fig:workflow}).} 
We randomly check around 300 figures from previous step, and find that around 80\% of them are compound figures, {i.e.} multiple pannels.
Here, we train a detector to break the compound figures into subfigures,
specifically, we use a ResNet-34-based DETR model~\cite{carion2020detr} with 4 encoder layers, 4 decoder layers, and 32 learnable queries.
The detector is trained on the MedICaT subset~\cite{subramanian2020medicat}, that has manually separated 2,069 compound figures and corresponding captions, which will be referred as MedICaTSub. 
For hyper-parameter tuning, we split the dataset into train and test set with a ratio of 3:1, our model obtains mAP@0.5 of 0.94 on the test set. 
To balance precision~(0.93) and recall~(0.94), we take the confidence threshold as 0.7.
After breaking the compound figures, non-medical subfigures like charts may be mixed with medical ones, we therefore filter the derived subfigures with the classification model~(repeat the first stage). 
Till here, we have obtained 1,646,592 subfigures from 378,717 compound figures, the number of captions is same as compound figures, thus each caption is assigned to an average of 4.3 subfigures at the moment.
We termed this dataset as a \textbf{\dataset{}-Beta} version.




\vspace{3pt}
\noindent \textbf{Subcaption Separation~$\&$~Alignment~(Step 5\&6 in Fig.~\ref{fig:workflow}).}
To further align subfigure to its corresponding part within the full caption, 
{\em i.e.}, subcaption, we need to break the captions into subcaptions first.
Here, we apply the caption distributor provided in~\cite{schwenker2021exsclaim} to all the captions in \dataset{} Beta. \textbf{Note that}, the tool sometimes fail to separate the caption, as a result, we get 898,467 subcaptions from 239,905 separable captions, corresponding to 1,003,911 subfigures. 
To align each subfigure to the most related subcaption, 
we pretrain a CLIP-style model on \textbf{\dataset{}-Beta}~(the training detail will be described in Sect.~\ref{sec:pmc-clip}), then finetune it on MedICaTSub. Specifically, for each subfigure, we finetune the pretrained model to match its subcaption with contrastive learning.
We split the subset into train and test by 3:1, 
and the finetuned model achieves alignment accuracy=73\% on test set. 
We finally align 1,003,911 subfigure-subcaption pairs. 
Along with the remaining 642,681 subfigure-caption pairs, 
we termed this dataset as\textbf{ \dataset{}}.
We consequently pretrain the \textbf{\model{}} on it.


\vspace{3pt}
\noindent \textbf{Discussion.} 
Some pioneering works~\cite{pelka2018roco,subramanian2020medicat} have explored publicly available scientific documents to alleviate the data scarcity in the biomedical domain. As they provide valuable datasets for the community, they suffer from some limitations. Our work aims to improve the collection pipeline for a more massive, diverse, and accurate dataset:
\textit{First},
\dataset{} covers a wider range of papers~(2,478,267) than ROCO~\cite{pelka2018roco}(1,828,575) and MedICaT~\cite{subramanian2020medicat}(131,410), and thus enlarge the dataset(1.6M). 
\textit{Second}, different from ROCO~\cite{pelka2018roco}, we maintain the non-radiology images, which makes \dataset{} contain more diverse biomedical data. 
We present the quantitative comparison in Fig.~\ref{fig:radar statistics}. \textit{Third}, to the best of our knowledge, we are the first to integrate subfigures separation, subcaptions separation and the alignment into the data collection pipeline, which explicitly enlarges our dataset~(approximately 8 times of MedICaT and 20 times of ROCO) while reducing the noise as much as possible. 

\subsection{Dataset Overview}
\label{sec:dataset_overview}

\begin{figure}[!t]
  \subfigure[Diagnostic procedure.]{
  \begin{minipage}[t]{0.49\linewidth}
    \centering
    \includegraphics[width=\textwidth]{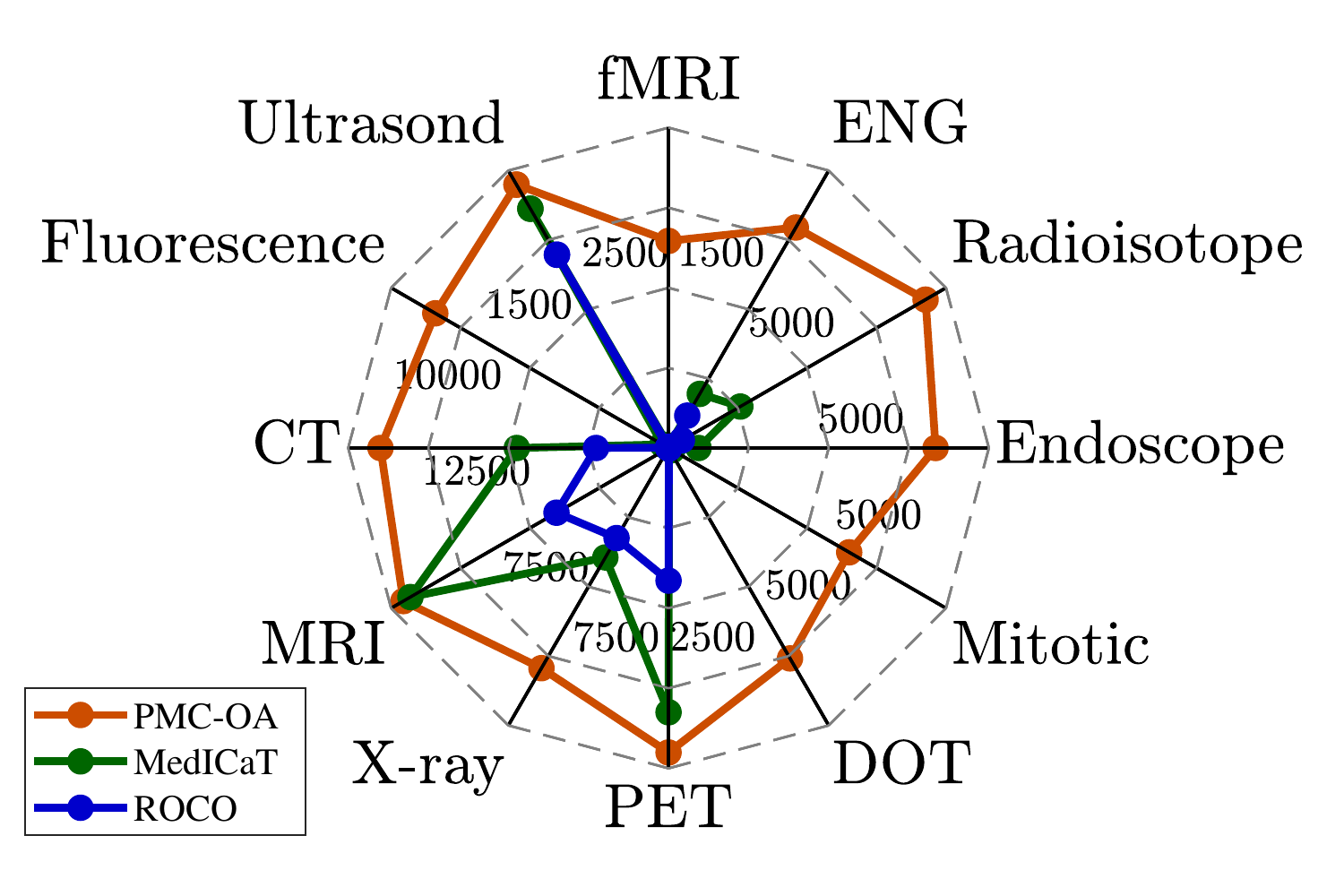}
    \label{fig:side:a}
  \end{minipage}
  }
  \subfigure[Disease and findings.]{
  \begin{minipage}[t]{0.49\linewidth}
    \centering
    \includegraphics[width=\textwidth]{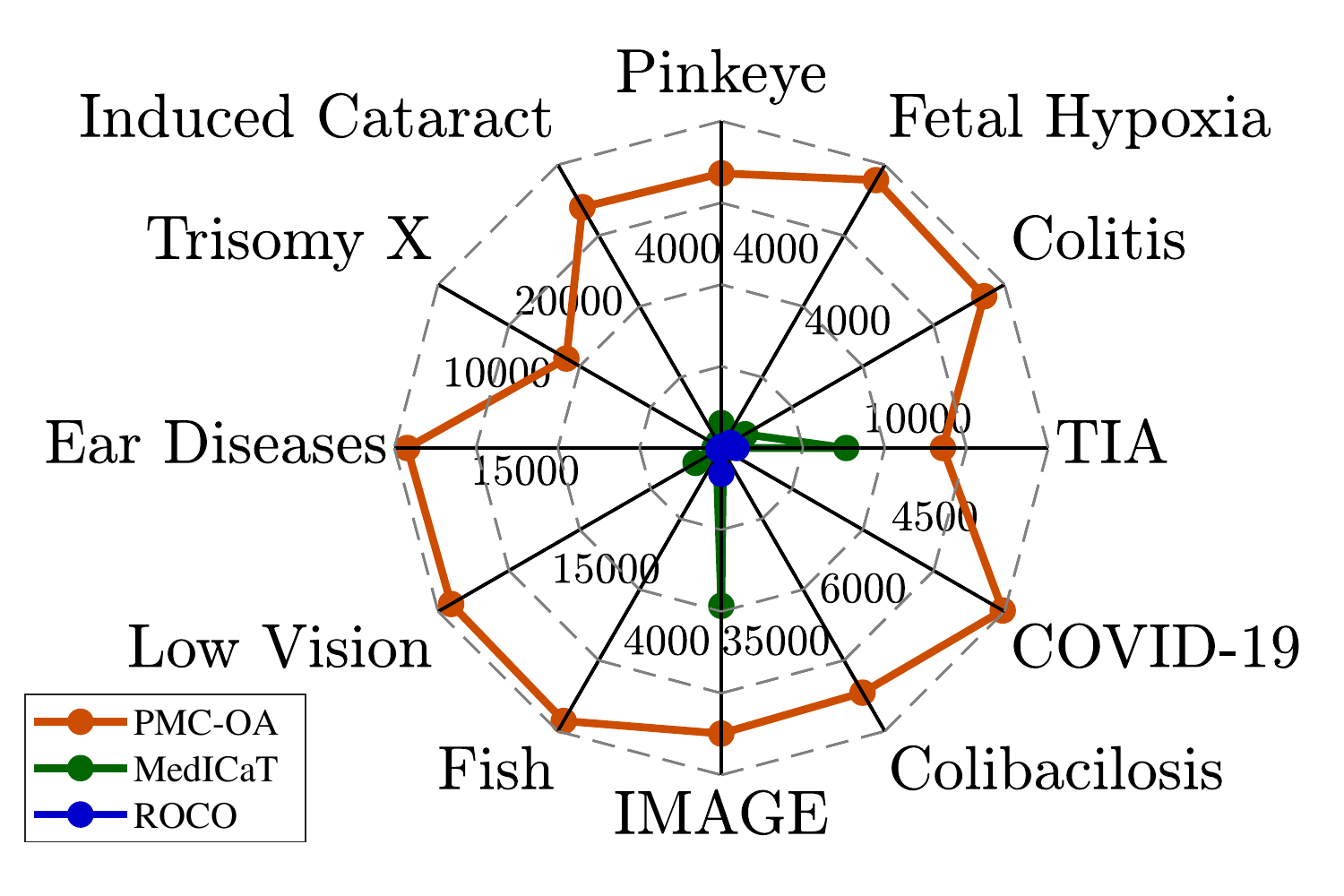}
    \label{fig:side:b}
  \end{minipage}
  }
  
  \subfigure[Disease distribution in \dataset{}.]{
    \begin{minipage}[t]{0.49\linewidth}
	\centering
        \includegraphics[width=\textwidth]{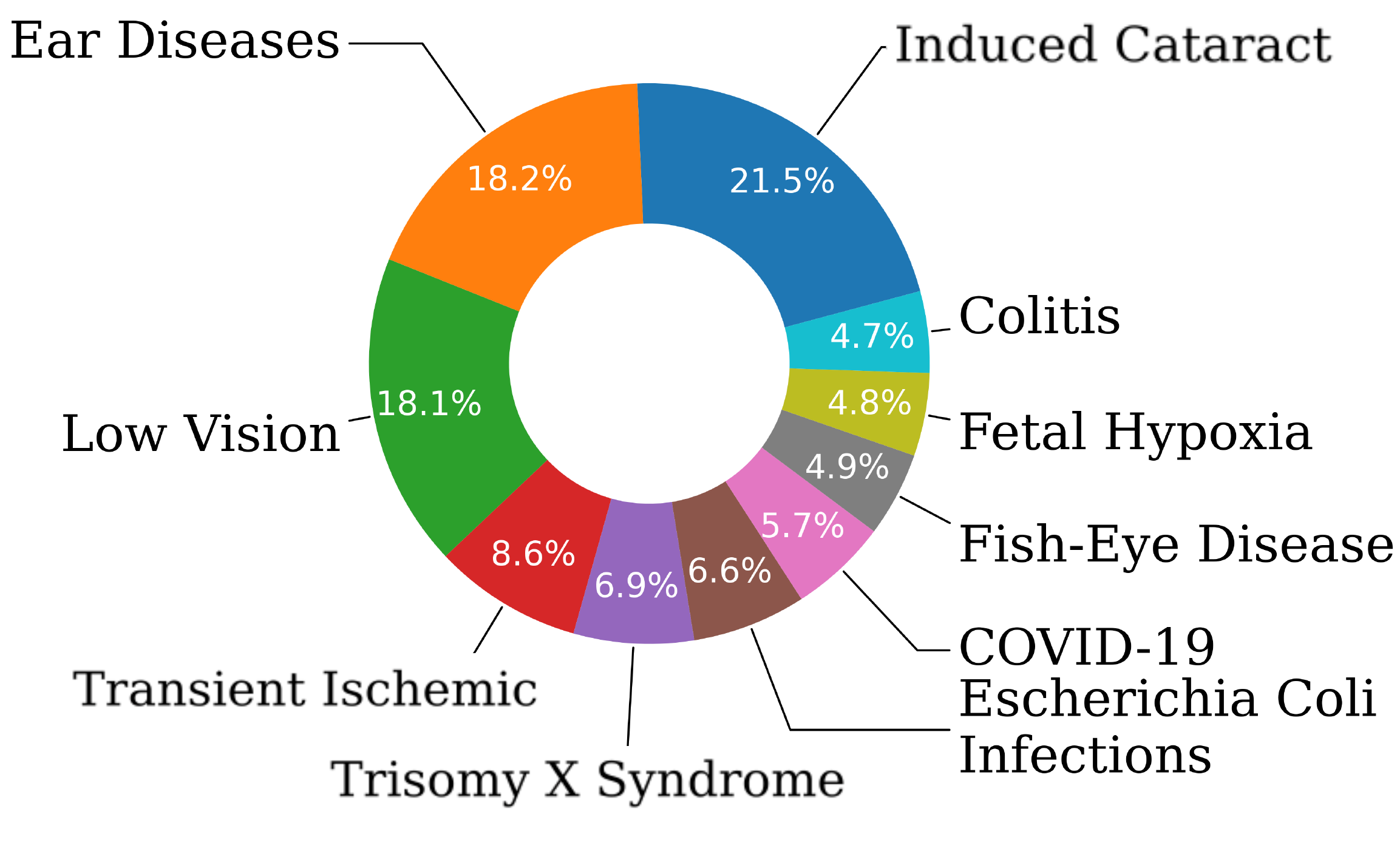}
        \label{fig:disease}
		\end{minipage}
    }
  \subfigure[Patients' age \& gender in \dataset{}.]{
        \begin{minipage}[t]{0.49\linewidth}
		\centering
		\includegraphics[width=\textwidth]{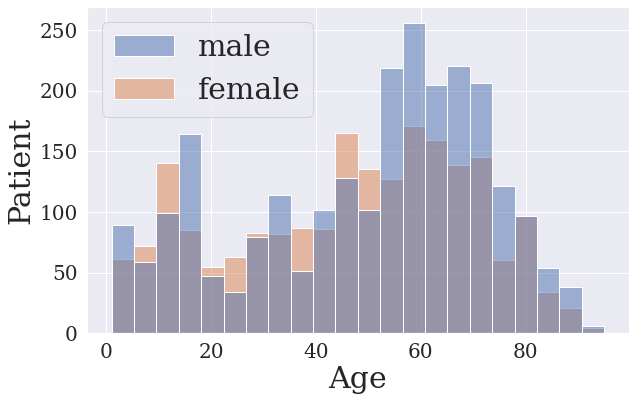}
        \label{fig:ages}
		\end{minipage}
    }
\caption{
Statistical overview of \dataset.
}
%
\label{fig:radar statistics}
\end{figure}

In this section, we provide a brief statistical overview of the collected dataset \dataset{} from three different perspectives,
{\em i.e.}, the diagnostic procedure, diseases and findings, and fairness.

\vspace{3pt}
\noindent \textbf{Diagnostic Procedure.}
As shown in Fig.~\ref{fig:side:a}, \dataset~covers a wide range of diagnostic procedures, spanning from common~({\em CT, MRI, X-ray}) to rare ones~({\em mitotic figure}), which is much diverse than before.

\vspace{3pt}
\noindent \textbf{Disease and Findings.}
In \dataset, diseases are given in the free-form text, allowing for elaborate identification and analysis. For instance, eye diseases can be further categorized into Cataracts, Conjunctivitis, Macular degeneration, fish eye disease, etc.
Fig.~\ref{fig:side:b} illustrates the frequently used words in diagnosis and Fig.~\ref{fig:disease} shows their distribution.

\vspace{3pt}
\noindent \textbf{Fairness.}
We provide the sex-ratio across ages in Fig.~\ref{fig:ages},
as we can see \dataset is approximately gender-balanced, with 54\% males.


\vspace{3pt}
\noindent \textbf{Discussion.} 
The detailed dataset statistics indicate the superiority of PMC-OA from three aspects: (i) diagnostic-procedure diversity; (ii) disease covering; (iii)  polpulation fairness. {\em First}, until now, the most widely-used text-image dataset is MIMIC-CXR\cite{johnson2019mimic} which contains only chest X-ray images, greatly limiting the potential of VLP methods and our dataset can compensate this. {\em Second}, diagnosis is always a crucial procedure in clinical, and the wide disease coverage in our dataset supports learning the shared patterns of diseases, promoting accurate auto-diagnosis. {\em Third}, the fairness on population ensures our dataset sightly suffers from patient characteristic bias, thus providing greater cross-center generalize ability.

%% file: content/03-Method.tex
\section{Visual-language Pre-training}
\label{sec:pmc-clip}

With our constructed image-caption dataset, 
we further train a visual-language model, termed as \model{} as shown in Fig.~\ref{fig:workflow}~(bottom). We describe the architecture in Sec.~\ref{sec:model} and introduce the training objectives in Sec.~\ref{sec:objective}

\subsection{Architecture}
\label{sec:model}

Given a dataset with $N$ image-caption pairs, 
{\em i.e.}, $\mathcal{D} = \{ (\mathcal{I}_1, \mathcal{T}_1), \cdots, (\mathcal{I}_N, \mathcal{T}_N) \}$, where $\mathcal{I}_i \in \mathbb{R}^{H \times W \times C}$ represents images, $H,W,C$ are height, width, channel, 
and $\mathcal{T}_i$ represents the paired text.
We aim to train a CLIP-style visual-language model with an image encoder $\mathrm{\Phi}_\text{visual}$ and a text encoder~$\mathrm{\Phi}_\text{text}$.

In detail, given a specific image-caption pair $(\mathcal{I}, \mathcal{T})$
we encode it separately with a ResNet-based $\mathrm{\Phi}_\text{visual}$ and a BERT-based $\mathrm{\Phi}_\text{text}$, the embedding dimension is denoted as $d$ and the text token length as $l$:
\begin{align}
   &\bm{v} = \mathrm{\Phi}_\text{visual}(\mathcal{I}) \in \mathbb{R}^d, \\
    & \bm{T} = \mathrm{\Phi}_\text{text}(\mathcal{T}) \in \mathbb{R}^{l\times d}, \text{  } \bm{t} = \bm{T}_0 \in \mathbb{R}^d,
\end{align}
where $\bm{v}$ represents the embedding for the whole image, 
$\bm{T}$ refers to the sentence embedding, 
and $\bm{t}$ denotes the embedding for [CLS] token.

%
%

\subsection{Training Objectives}
\label{sec:objective}

In this section, we train the visual-language model with two objectives, \emph{i.e.},
Image-Text Contrastive learning and Masked Language Modeling.

\vspace{5pt}
\noindent \textbf{Image-Text Contrastive Learning~(ITC).} 
We implement ITC loss following CLIP~\cite{radford2021learning},
that aims to match the corresponding visual and text representations from one sample. 
In detail,  denote batch size as $b$, we calculate the softmax-normalized cross-modality dot product similarity between the current visual/text embedding~($\bm{v}$ / $\bm{t}$) and all samples within the batch, 
termed as $p^{\text{i2t}}, p^{\text{t2i}} \in \mathbb{R}^{b}$, 
and the final ITC loss is:
\begin{equation}
    L_{\text{ITC}} = \mathbb{E}_{(\mathcal{I}, \mathcal{T})\sim \mathcal{D}} \big[ \text{CE}( y^{\text{i2t}}, p^{\text{i2t}} ) + \text{CE}( y^{\text{t2i}}, p^{\text{t2i}} ) \big],
\end{equation}
where $ y^{\text{i2t}}, y^{\text{t2i}}$ refer to one-hot matching labels, $\text{CE}$ refers to InfoNCE loss \cite{oord2018representation}.

\vspace{5pt}
\noindent \textbf{Masked Language Modeling~(MLM).}
We implement MLM loss following BERT~\cite{devlin2018bert}.
The network is trained to reconstruct the masked tokens from context contents and visual cues. 
We randomly mask the word in texts with a probability of $15\%$ and replace it with Token [MASK]. We concate the image embedding $\bm{v}$ with the sentence embedding $\bm{T}$, input it into a self-attention transformer-based fusion module~$\mathrm{\Phi}_\text{fusion}$, and get the prediction for the masked token at the corresponding position in the output sequence, termed as $p^{\text{mask}} = \mathrm{\Phi}_\text{fusion}(\bm{v},\bm{T})$. Let $y^{\text{mask}}$ denote the ground truth, and the MLM loss is:
\begin{equation}
    L_{\text{MLM}} = \mathbb{E}_{(\mathcal{I}, \mathcal{T})\sim \mathcal{D}} \big[ \text{CE}(y^{\text{mask}}, p^{\text{mask}}) \big]
\end{equation}

\vspace{5pt}
\noindent \textbf{Total Training Loss. } 
The final loss is the weighted sum of the above two:
\begin{equation}
    L = L_{\text{ITC}} + \lambda L_{\text{MLM}},
\end{equation}
where $\lambda$ is a hyper-parameter deciding the weight of $L_{\text{MLM}}$, set as $0.5$ by default.

%% file: content/04-Experiment.tex
\section{Experiment Settings}
Here, we start by introducing the compared datasets in Sec.~\ref{sec:pretraining_dataset}, describe the downstream tasks in Sec.~\ref{sec:exp_downstream}, cover
the implementation details in Sec.~\ref{sec:implementation}.
\vspace{-10pt}

\subsection{Pre-training Datasets}
\label{sec:pretraining_dataset}

\noindent \textbf{ROCO} \cite{pelka2018roco} is a image-caption dataset collected from PubMed~\cite{roberts2001pubmed}. It filters out all the compound or non-radiological images, and consists of 81K samples. 

\vspace{5pt}
\noindent \textbf{MedICaT} \cite{subramanian2020medicat} extends ROCO to 217K samples~(image-caption pairs), 
however, 75\% of its figures are compound ones, 
{\em i.e.} one figure with multiple subfigures. 

\vspace{5pt}
\noindent \textbf{MIMIC-CXR} \cite{johnson2019mimic} is the largest chest X-ray dataset, 
containing 377,110 samples~(image-report pairs). 
Each image is paired with a clinical report describing findings from doctors. 


\subsection{Downstream Tasks}
\label{sec:exp_downstream}

\noindent \textbf{Image-Text Retrieval~(ITR). } 
ITR contains both image-to-text(I2T) and text-to-image(T2I) retrieval. 
We train \model{} on different datasets, and evaluate on the ROCO testset.
\textbf{Note that}, 
we have  explicitly conducted duplication between our data and ROCO, 
our reported results thus resembles {\em zero-shot} evaluation.
Following previous works~\cite{subramanian2020medicat,chen2022multi,chen2022align}, 
we sample 2,000 image-text pairs from ROCO's testset for evaluation.

\vspace{3pt}
\noindent \textbf{Classification. } 
We finetune the model for different downstream tasks that focus on image classification.
Spcifically, MedMINIST \cite{yang2023medmnist} contains 12 tasks in total, and it covers primary data modalities in biomedical images, including Colon Pathology, Dermatoscope, Retinal OCT, etc.

\vspace{3pt}
\noindent \textbf{Visual Question Answering~(VQA). } We evaluate on the official dataset split of VQA-RAD \cite{lau2018dataset}, SLAKE \cite{liu2021slake},
where SLAKE is composed of 642 images and 14,028 questions and VQA-RAD contains 315 images and 3,515 questions. The questions in VQA-RAD and Slake are categorized as close-ended if answer choices are limited, otherwise open-ended.
While adapting our PMC-CLIP to VQA task, we maintain most of the pre-trained parameters, including visual encoder, text encoder, and fusion.
Specifically, the image and question are fed into PMC-CLIP, 
the output embedding from fusion module is then used to compute similarity between each of the answer candidates~(also encoded with text encoder).
To strive for better adaptation, 
we use 10 learnable answer and question prompt vectors in text encoder repectively.


%

\vspace{3pt}
\noindent \textbf{Metrics.} 
We use Accuracy and AUC are for classification, Recall@K(K=1,5,10) for retrieval and Accuracy for VQA.

\subsection{Implementation Details}
\label{sec:implementation}
For the visual and text encoders, we adopt ResNet50~\cite{he2016deep} and PubmedBERT~\cite{gu2021domain}.
And we use 4 transformer layers for the fusion module.
For input data, we resize each image to $224 \times 224$.
During pre-training, our text encoder is initialized from PubmedBERT, while the vision encoder and fusion module are trained from scratch.
We use AdamW~\cite{loshchilov2017decoupled} optimizer with $lr =1\times 10^{-4}$.
We train on GeForce RTX 3090 GPUs with batch size 128 for 100 epochs.
The first 10 epochs are set for warming up.

%% file: content/05-Result.tex
\section{Result}

We conduct experiments to validate our proposed dataset, 
and the effectiveness of models trained on it.
In Sec.~\ref{sec:dataset_ablation}, 
we first compare with existing large-scale biomedical datasets on the image-text retrieval task to demonstrate the superiority of PMC-OA. 
In Sec.~\ref{sec:2}, we finetune the model~(pre-trained on PMC-OA) across three different downstream tasks, namely, retrieval, classification, and visual question answering. And we also perform a thorough empirical study of the pretraining objectives and the model architectures in Sec.~\ref{sec:3}.
\textbf{Note that}, for all experiments, unless specified otherwise,
we use the default setting: ResNet50 for image encoder, and pre-train with both ITC and MLM objectives.


\subsection{PMC-OA surpasses SOTA large-scale biomedical dataset}
\label{sec:dataset_ablation}

As shown in Tab \ref{tab:ablation_pretrain}, 
we pre-train \model~on different datasets and evaluate the retrieval on ROCO test set.
The performance is largely improved while simply switching to our dataset, 
confirming the significance of it.

\begin{table*}[!htb]
    \centering
    \scriptsize
    \setlength{\tabcolsep}{4pt} 
    \caption{Ablation studies on pre-training dataset.}
    \begin{tabular}{lcc|ccc|ccc}
        \toprule
        \multirow{2}{*}{Methods} & \multirow{2}{*}{Pretrain Data} & \multirow{2}{*}{DataSize} & \multicolumn{3}{c|}{I2T} & \multicolumn{3}{c}{T2I} \\
        ~ & ~ & ~ & R@1 & R@5 & R@10 & R@1 & R@5 & R@10 \\
        \midrule
        \model~& \scriptsize ROCO & 173 K & 12.30 & 35.28 & 46.52 & 13.36 & 35.84 & 47.38 \\
        \model~& \scriptsize MedICaT & 173 K & 12.30 & 35.28 & 46.52 & 13.36 & 35.84 & 47.38 \\
        \model~& \scriptsize \datasetbeta{} & 1.6 M & \cellcolor{lightergray}30.42 & \cellcolor{lightergray}59.11 & \cellcolor{lightergray}70.16 & \cellcolor{lightergray}27.92 & \cellcolor{lightergray}55.99 & \cellcolor{lightergray}66.35 \\
        \model{} & \scriptsize \dataset{} & 1.6 M & \cellcolor{lightgray}31.41 & \cellcolor{lightgray}61.15 & \cellcolor{lightgray}71.88 & \cellcolor{lightgray}28.02 & \cellcolor{lightgray}58.33 & \cellcolor{lightgray}69.69  \\
        \midrule
    \end{tabular}
    \label{tab:ablation_pretrain}
\end{table*}

\subsection{PMC-CLIP achieves SOTA across downstream tasks}
\label{sec:2}

To evaluate the learnt representation in PMC-CLIP, we compare it with several state-of-the-art approaches across various downstream tasks, 
including image-text retrieval, image classification, and visual question answering.

\vspace{3pt}
\noindent \textbf{Image-Text Retrieval. }
As shown in Tab.~\ref{tab:Zeroshot-ROCO}, 
we report a state-of-the-art result on image-text retrieval.
On I2T Rank@10, \model~outperforms previous state-of-the-art by 8.1\%.
It is worth mentioning that, the training set of ROCO has been used during pretraining in M3AE~\cite{chen2022multi}, ARL~\cite{chen2022align}.
While our dataset does not contain data from ROCO.


\vspace{-5pt}
\input{content/tab/retrieval.tex}

\vspace{-5pt}

\noindent \textbf{Image Classification. }
\noindent To demonstrate the excellent transferability of \model, 
we finetune it on MedMNIST and compare it with SOTA methods, 
{\em i.e.,}, DWT-CV~\cite{cheng2022dwt} and SADAE~\cite{ge2022self}.
We present the results of 3 of 12 sub-tests here, 
and the full results can be found in the supplementary material.
As shown in Tab.~\ref{tab:MedMNIST},
\model~obtains consistently higher results, 
and it is notable that finetuning from \model~achieves significant performance gains compared with training from scratch with ResNet.

\input{content/tab/classification.tex}

\vspace{3pt}
\noindent \textbf{Visual Question Answering. }
\noindent VQA requires model to learn finer grain visual and language representations. As Table \ref{tab:vqa} shows, we surpass SOTA method M3AE in 5 out of 6 results.
\input{content/tab/vqa.tex}
\vspace{-20pt}

\subsection{Ablation Study}
\label{sec:3}


For the effectiveness illustration of the pretraining objectives ({\em ITC, MLM}), 
we evaluate \model{} pre-trained on \dataset{} for ablation studies.
To validate the necessity of fusion module, 
we compare MLM using text-only context~({\em MLM-T}), 
and MLM with visual cues~({\em MLM-V}).

As shown in Tab.~\ref{tab:ablation}, we can draw the following observations:
{\em First}, ITC objective is essential for pretraining, and contributes most of the performance~(ID 1).
%
{\em Second}, MLM using only text context works as a regularization term (ID 2). 
{\em Third}, With incorporation of visual features, the model learns finer grain correlation between image-caption pairs, and achieve the best results (ID 3).

\vspace{-8pt}

\input{content/tab/ablation}
\vspace{-20pt}

%% file: content/tab/retrieval.tex
\begin{table*}[htpb]
    \centering
    \scriptsize
    \setlength{\tabcolsep}{4pt} 
    \renewcommand{\arraystretch}{1} 
    \caption{Zero-shot Image-Text Retrieval on ROCO. Dark and light grey colors highlight the top and second best results on each metric.
    }
    \vspace{-5pt}
    \begin{tabular}{lcc|ccc|ccc}
    \toprule
        \multirow{2}{*}{Methods} & \multirow{2}{*}{Pretrain Data} & \multirow{2}{*}{DataSize} & \multicolumn{3}{c|}{I2T} & \multicolumn{3}{c}{T2I} \\
        ~ & ~ & ~ & R@1 & R@5 & R@10 & R@1 & R@5 & R@10 \\
        \toprule
        ViLT \cite{kim2021vilt} & COCO, VG, SBU, GCC & 4.1M & 11.90 & 31.90 & 43.20 & 9.75 & 28.95 & 41.40 \\
        METER \cite{dou2022empirical} & COCO, VG, SBU, GCC & 4.1M & 14.45 & 33.30 & 45.10 & 11.30 & 27.25 & 39.60 \\
        M3AE \cite{chen2022multi} & \scriptsize ROCO, MedICaT & 233 K & 19.10 & 45.60 & 61.20 & 19.05 & 47.75 & 61.35 \\
        ARL \cite{chen2022align} & \scriptsize ROCO, MedICaT, CXR & 233 K & \cellcolor{lightergray}23.45 & \cellcolor{lightergray}50.60 & \cellcolor{lightergray}62.05 & \cellcolor{lightergray}23.50 & \cellcolor{lightergray}49.05 & \cellcolor{lightergray}63.00 \\
        \midrule
        PMC-CLIP & \scriptsize PMC-OA & 1.6 M & \cellcolor{lightgray}31.41 & \cellcolor{lightgray}61.15 & \cellcolor{lightgray}71.88 & \cellcolor{lightgray}28.02 & \cellcolor{lightgray}58.33 & \cellcolor{lightgray}69.69  \\
        \bottomrule
    \end{tabular}
    \label{tab:Zeroshot-ROCO}
\end{table*}

%% file: content/tab/classification.tex
\begin{table*}[htpb]
    \centering
     \scriptsize
    \setlength{\tabcolsep}{10pt} 
    \caption{Classification results on MedMNIST. Dark and light grey colors highlight the top and second best results on each metric.}
    \vspace{-5pt}
    \begin{tabular}{l|cc|cc|cc}
    \toprule
        Dataset & \multicolumn{2}{c|}{PneumoniaMNIST} & \multicolumn{2}{c|}{BreastMNIST} & \multicolumn{2}{c}{DermaMNIST} \\
        Method & AUC$\uparrow$ & ACC$\uparrow$ & AUC$\uparrow$ & ACC$\uparrow$ & AUC$\uparrow$ & ACC$\uparrow$ \\
        \toprule
        ResNet50~\cite{he2016deep} & 96.20 & 88.40 & \cellcolor{lightergray}91.90 & 86.60 & 91.20 & 73.10 \\
        DWT-CV \cite{cheng2022dwt} & 95.69 & 88.67 & 89.77 & 85.68 & 91.67 & 74.75 \\ 
        SADAE \cite{ge2022self} & \cellcolor{lightergray}98.30 & \cellcolor{lightergray}91.80 & 91.50 & \cellcolor{lightergray}87.80 & \cellcolor{lightergray}92.70 & \cellcolor{lightergray}75.90 \\
        \midrule
        PMC-CLIP & \cellcolor{lightgray} 99.02 & \cellcolor{lightgray}95.35 & \cellcolor{lightgray}94.56 & \cellcolor{lightgray}91.35 & \cellcolor{lightgray}93.41 & \cellcolor{lightgray}79.80 \\
     \bottomrule
    \end{tabular}
    \label{tab:MedMNIST}
\end{table*}

%% file: content/tab/vqa.tex

\begin{table*}[htpb]
    \centering
    \scriptsize
    \setlength{\tabcolsep}{6pt} 
    \caption{VQA results on VQA-RAD and Slake}
    \begin{tabular}{c|ccc|ccc}
    \toprule
        \multirow{2}{*}{Methods} & \multicolumn{3}{c|}{VQA-RAD} & \multicolumn{3}{c}{Slake} \\
         & Open & Closed & Overall & Open & Closed & Overall \\
        \midrule
        MEVF-BAN \cite{nguyen2019overcoming} & 49.20 & 77.20 & 66.10 & 77.80 & 79.80 & 78.60 \\
        CPRD-BAN \cite{liu2021contrastive} & 52.50 & 77.90 & 67.80 & 79.50 & 83.40 & 81.10 \\
        M3AE \cite{chen2022multi} & \cellcolor{lightgray}67.23 & \cellcolor{lightergray}83.46 & \cellcolor{lightergray}77.01 & \cellcolor{lightergray}80.31 & \cellcolor{lightergray}87.82 & \cellcolor{lightergray}83.25 \\
        \midrule
        PMC-CLIP & \cellcolor{lightergray}67.00 & \cellcolor{lightgray}84.00 & \cellcolor{lightgray}77.60 & \cellcolor{lightgray}81.90 & \cellcolor{lightgray}88.00 & \cellcolor{lightgray}84.30 \\
        \bottomrule
    \end{tabular}
    \label{tab:vqa}
\end{table*}

%% file: content/tab/ablation.tex
\begin{table*}[htpb]
    \centering
    \scriptsize
    \setlength{\tabcolsep}{2pt} 
    \caption{Ablation studies of pre-training objectives}
    \begin{tabular}{cccc|ccc|cc|cc|cc}
    \toprule
        \multirow{2}{*}{ID} & \multicolumn{3}{c|}{Methods} & \multicolumn{3}{c|}{ROCO} & \multicolumn{2}{c|}{\scriptsize PneumoniaMNIST} & \multicolumn{2}{c|}{\scriptsize BreastMNIST} & \multicolumn{2}{c}{\scriptsize DermaMNIST} \\
        ~ & ITC & MLM-T & MLM-V & R@1 & R@5 & R@10 & AUC & ACC & AUC & ACC & AUC & ACC \\
        \midrule
        1 & \usym{2713} & & & 25.53 & 53.48 & 64.64 & 98.07 & 93.56 & 93.04 & 89.74 & \cellcolor{lightergray}92.77 & 77.91 \\
        2 & \usym{2713} & \usym{2713} & ~ & \cellcolor{lightergray}29.17 & \cellcolor{lightergray}57.55 & \cellcolor{lightergray}68.26 & \cellcolor{lightergray}98.53 & \cellcolor{lightergray}94.31 & \cellcolor{lightergray}94.39 & \cellcolor{lightergray}90.71 & 92.76 & \cellcolor{lightergray}78.00 \\
        3 & \usym{2713} & ~ & \usym{2713} & \cellcolor{lightgray}29.72 & \cellcolor{lightgray}59.74 & \cellcolor{lightgray}70.79 & \cellcolor{lightgray}99.02 & \cellcolor{lightgray}95.35 & \cellcolor{lightgray}94.56 & \cellcolor{lightgray}91.35 & \cellcolor{lightgray}93.41 & \cellcolor{lightgray}79.80 \\
        \midrule
    \end{tabular}
    \label{tab:ablation}
\end{table*}

%% file: content/06-Conclusion.tex
\section{Conclusion}

In this paper, we present a large-scale dataset in biomedical domain, named \dataset{},
by collecting image-caption pairs from abundant scientific documents.
We train a CLIP-style model on on \dataset{}, termed as \model{},
it achieves SOTA performance across various downstream biomedical tasks, including image-text retrieval, image classification, visual question answering.
With the automatic collection pipeline, we believe the dataset will benefit the research community, fostering the development of foundation models in biomedical domain.